\documentclass[10pt]{article}
\usepackage[margin=1in]{geometry}
\usepackage{pgfplots}
\pgfplotsset{compat=1.17}
\usepackage{booktabs}
\usepackage{amsmath}
\usepackage{hyperref}
\usepackage{natbib}
\usepackage{authblk}

\title{\bfseries Minimum Description Length Unsupervisedly Separates
Law-Grade from Statistical-Grade Knowledge in a Self-Improving
Science Agent\\[2pt]
\large\bfseries --- and Reveals 86\% of Its Own Discoveries Are
Statistical}
\author[1]{Wenbin Mao}
\author[1]{Pei Zou}
\affil[1]{Artificial Intelligence Laboratory, Hooman Biotechnology
(Beijing) Co., Ltd.\\ \texttt{research.ai@xthinktank.com}}
\date{October 23, 2026}

\begin{document}
\maketitle

\begin{abstract}
AI systems that discover scientific relations now generate candidate
``laws'' at scale, but the bottleneck has shifted from generation to
\emph{vetting}: which candidates deserve belief? We show that a
deployed self-improving science agent---one that has admitted 252
relations through held-out prediction courts---already computes the
quantity that answers this question: the Minimum Description Length
compression ratio used for model selection doubles, when read under
discrete accounting, as an unsupervised \emph{epistemic grader}.
The compression-ranked top band of the agent's ledger recovers exactly
the canonical laws among its discoveries (vis-viva across seven
planets, Balmer/Rydberg) plus cross-measurement identities---
enrichment $p=3.6\times10^{-12}$ under a hypergeometric test---with
zero counterexamples and physically coherent band structure.
The same lens retroactively reveals that 86\% of the ledger compresses
nothing ($\mathrm{cr}\approx 1$): the agent had been recording
statistical regularities, including a learned predictor with held-out
$R^2=0.541$ admitted through the prediction gate, under a narrative of
compression as the unified value of knowledge---a narrative that
was never true for them.
We formalize the resulting \emph{dual-axis bookkeeping}---prediction
value ($R^2$) and compression value (cr) as nearly decoupled
axes of knowledge---demonstrate its robustness to quantization, and register
forward falsifiable predictions. The episode is, to our knowledge, the
first measured instance of a discovery agent auditing its own knowledge
ledger by an internal measure and finding itself over-claiming.
\end{abstract}

\section{Introduction}

AI systems that discover scientific relations---symbolic regression
\citep{schmidt2009}, concept-driven law discovery \citep{ainewton2025},
and LLM-based pipelines---now generate candidate ``laws'' at scale.
The binding constraint has quietly shifted from \emph{generation} to
\emph{vetting}: which of hundreds of admitted candidates deserve
belief? The concern is now institutional \citep{cornelio2026},
autonomous-research systems draw systematic critique for exactly
this gap \citep{beel2025}, and the symbolic-regression literature
still grades discovery almost exclusively by recovery of known
ground-truth formulas \citep{matsubara2022}---a retreat made
necessary by the absence of any grader that works without ground
truth. Human curation does not scale with the output of a machine
that never sleeps, and the systems themselves carry no instrument
that grades the epistemic status of what they find. A discovery
system that cannot tell a law from a strong correlation will bank
both under the same heading---and, given enough time, mistake its
own statistics for physics.

This paper reports that such an instrument already exists inside at
least one deployed system, because the quantity used for model
selection \emph{is} the grader when read correctly. The agent under
study is a self-improving physical-science runtime that has admitted
252 relations through held-out prediction courts, each scored for
predictive $R^2$ and for a Minimum Description Length (MDL)
compression ratio \citep{rissanen1978,grunwald2007}. Under the
discrete two-part accounting defined in \S\ref{sec:system}, the
compression gain is effectively zero for statistical relations (the
ratio sits at parity, $\mathrm{cr}\approx1$) and explodes only as
residuals approach determinism---a \emph{law cliff}
(\S\ref{sec:cliff}). Auditing the ledger through this lens yields
three findings:

\begin{enumerate}
\item \textbf{Unsupervised law recovery}
(\S\ref{sec:recovery}). The compression-ranked top band of the
ledger consists exactly of the canonical laws and cross-measurement
identities among the agent's discoveries---the vis-viva relation
across seven planets from JPL Horizons ephemeris, Balmer/Rydberg
lines from NIST measurements, double-measurement and cell-volume
identities---with enrichment $p=3.6\times10^{-12}$ under a
hypergeometric test and zero counterexamples.
\item \textbf{A measured self-deception event} (\S\ref{sec:audit}).
The same lens shows 86\% of the ledger (216 of 252 relations)
compresses nothing ($\mathrm{cr}\approx 1$): the agent had been
recording statistical regularities---including a learned predictor
with held-out $R^2=0.541$ that passed the prediction gate---under a narrative of
compression as the unified value of knowledge---a narrative that
was never true for them. The
narrative was written into the system's written charter before the audit;
the audit falsified it.
\item \textbf{Dual-axis bookkeeping} (\S\ref{sec:audit}). Prediction
value and compression value are nearly decoupled axes: $R^2=0.9$ still
yields $\mathrm{cr}=1.06$. A discovery agent that optimizes only
prediction will accumulate statistics and mistake them for laws. We
record the corrective---dual bookkeeping with explicit
grades---and register forward falsifiable predictions
(\S\ref{sec:prereg}) that make the grading claim testable rather
than retrospective.
\end{enumerate}

To our knowledge this is the first measured instance of a discovery
agent auditing its own knowledge ledger by an internal measure and
finding itself over-claiming. The math is not new---it is a Gaussian
shadow of Kolmogorov's structure function
\citep{vereshchagin2004}, and we claim no theoretical contribution.
What is new is the demonstration that in a running system the
measure separates laws from statistics at $10^{-12}$ significance,
that the failure mode it exposes is real and was present, and that
the fix is an accounting convention rather than a new mechanism.

\section{System and Accounting}\label{sec:system}

\paragraph{The agent.} The system is a self-improving
physical-science runtime (companion paper \citep{companion}).
Concretely, it continuously ingests public measurement archives
(planetary ephemerides, spectroscopy tables, crystallography
databases, climate and geophysical records), proposes parametric
relations of the form $y\approx f(x;\theta)$ between measured
variables of one dataset, and admits a relation only if it beats a
standard basis on a held-out split---gates vary by court family,
typically $R^2\ge0.5$ with a margin over the basis. Every admitted
relation (``symbol'') enters a persistent ledger.

\paragraph{What ``agent'' means here.} We use the word in its
classical sense \citep{russell2021}: an entity that perceives its
environment (measurement archives), acts autonomously on it
(admitting, rejecting, registering predictions), and maintains
persistent state toward its goal (a ledger of court-verified
knowledge). It is a long-running discovery runtime, not a trained
model, and not an LLM-based agent in the contemporary harness sense: no language model proposes,
admits, or grades any relation in the audited ledger---candidates are
parametric forms from symbolic generators, and admission is decided
by held-out statistics. A separate coding agent (Acknowledgments)
develops the runtime itself under human approval; that activity lies
outside the audited loop and is not claimed as the system's own
autonomy.

\paragraph{Discrete two-part accounting.} For a target $y$ of
variance $\sigma_y^2$ over $n$ samples, a model's residual
$\sigma_r^2$, and quantization $\delta=\sigma_y\cdot10^{-6}$, the
residual code is
$\frac{n}{2}\log_2\!\left(2\pi e\,\sigma_r^2/\delta^2\right)$ bits
and the baseline code substitutes $\sigma_y^2$. The description
length of the fitted form is added to the model side. The
compression ratio is
$\mathrm{cr}=\text{baseline}/\text{total}$.
Two properties matter here. First, $\mathrm{cr}$ is \emph{not} a
monotone re-labeling of $R^2$: it depends on
$\log(\sigma_r^2/\delta^2)$, so variance reduction trades against
the quantization floor. Second, exact relations---residuals at the
floor---saturate the residual code while the baseline grows with
$-\log\delta$, so $\mathrm{cr}\to\infty$ as quantization refines:
determinism, not fit quality, buys compression.

\paragraph{Provenance.} Every relation in the law band traces to
external measurement: planetary state vectors from JPL Horizons
ephemerides (positions and velocities; the vis-viva variables are
$r=|\mathbf{x}|$ and $v^2=|\dot{\mathbf{x}}|^2$), hydrogen line
wavelengths from NIST tables, crystallographic cell parameters from
COD. No law-band relation was generated from the formula it
rediscovers; the audit is not circular.

\section{The Law Cliff}\label{sec:cliff}

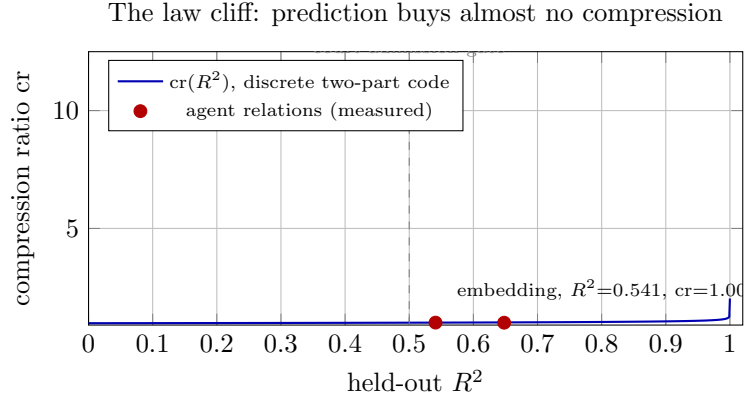
\begin{figure}[t]
\centering
\begin{tikzpicture}
\begin{axis}[width=0.62\textwidth, height=5.2cm,
  xlabel={held-out $R^2$}, ylabel={compression ratio $\mathrm{cr}$},
  title={The law cliff: prediction buys almost no compression},
  xmin=0, xmax=1.02, ymin=0.9, ymax=12.5, grid=major,
  legend pos=north west, legend style={font=\scriptsize}]
\addplot[thick, color=blue!70!black] table {fig2_cliff.dat};
\addlegendentry{$\mathrm{cr}(R^2)$, discrete two-part code}
\addplot[only marks, mark=*, mark size=2.4pt, color=red!70!black]
  table[x=r2, y=cr] {fig2_points.dat};
\addlegendentry{agent relations (measured)}
\node[font=\scriptsize, anchor=west] at (axis cs:0.56,2.4)
  {embedding, $R^2{=}0.541$, $\mathrm{cr}{=}1.004$};
\draw[dashed, gray] (axis cs:0.5,0.9) -- (axis cs:0.5,12.5);
\node[font=\scriptsize, gray, anchor=south] at (axis cs:0.5,11.9)
  {court admission gate};
\end{axis}
\end{tikzpicture}
\caption{Compression ratio as a function of held-out predictive
$R^2$ under the agent's discrete two-part accounting. The curve rises
logarithmically: even $R^2=0.9$ yields $\mathrm{cr}=1.06$; compression
explodes only as residuals approach the quantization floor, {i.e.},
deterministic law-hood. The two axes are nearly decoupled across the statistical
regime: a strong learned predictor (embedding) compresses
nothing.}
\label{fig:cliff}
\end{figure}

Figure~\ref{fig:cliff} plots $\mathrm{cr}(R^2)$ under the agent's
accounting. The curve is nearly flat over the statistical regime:
$R^2=0.541$ gives $\mathrm{cr}=1.004$ (the agent's best learned
predictor sits exactly on the curve); even $R^2=0.9$ gives only
$1.06$; $R^2=0.99$ gives $1.15$. Compression explodes only in the
last decimal places of residual variance---$R^2=1-10^{-12}$ gives
$\mathrm{cr}=8.7$---because only there do residuals reach the
quantization floor. The cliff sits at \emph{determinism}, not at
``good prediction.''

The split is structural, not an artifact of the quantization scale.
Refining $\delta$ from $\sigma_y\cdot10^{-3}$ to
$\sigma_y\cdot10^{-12}$ widens the law/statistic separation ratio
monotonically ($4.7\times\to16.5\times$): any exact relation
eventually dominates any statistical one, while statistical
relations converge to $\mathrm{cr}\approx1$ from below or above
depending on description cost. No choice of $\delta$ reorders the
two classes. By construction $\delta=\sigma_y\cdot10^{-6}$ makes
the baseline $n\cdot21.3$ bits regardless of the variable's scale:
$\mathrm{cr}$ is scale-free, and sample size enters only through
description amortization.

We claim no novelty for the mathematics; it is a Gaussian-code
instance of the structure/statistics separation formalized by
Kolmogorov's structure function and algorithmic statistics
\citep{vereshchagin2004}, and the noise-floor intuition traces to
Shannon. The contribution is that a running discovery system can
carry this separation as a live ledger column---and that, read
correctly, the column already answers the vetting question.

\section{Law Recovery: the Ledger Audit}\label{sec:recovery}

\begin{figure}[t]
\centering
\begin{minipage}[c]{0.60\textwidth}
\centering
\begin{tikzpicture}
\begin{axis}[width=\textwidth, height=6.2cm,
  xlabel={rank (all 252 admitted relations, cr-sorted)},
  ylabel={compression ratio $\mathrm{cr}$},
  xmin=0, xmax=253, ymin=0.95, ymax=4.8, grid=major,
  title={Compression accounting recovers the law band}]
\addplot[only marks, mark=square*, mark size=1.5pt,
  color=blue!55!black, forget plot]
  table[x=rank, y=cr] {fig1_all.dat};
\addplot[only marks, mark=triangle*, mark size=2.1pt,
  color=red!75!black]
  table[x=rank, y=cr] {fig1_law.dat};
\draw[dashed, gray] (axis cs:0,1.5) -- (axis cs:253,1.5);
\node[font=\scriptsize, gray, anchor=south] at (axis cs:200,1.55)
  {law-grade threshold};
\end{axis}
\end{tikzpicture}
\end{minipage}\hfill
\begin{minipage}[c]{0.36\textwidth}
\centering
\scriptsize
\begin{tabular}{rlc}
\toprule
rank & relation & $\mathrm{cr}$ \\
\midrule
1 & vis-viva (Mercury) & 4.62 \\
2 & Balmer--Rydberg & 3.04 \\
3 & double measurement & 3.02 \\
4 & cell-volume identity & 2.19 \\
5 & vis-viva (Mars) & 2.14 \\
6 & vis-viva (Venus) & 1.91 \\
7 & Rydberg $n^{-2}$ & 1.72 \\
8 & vis-viva (Jupiter) & 1.52 \\
9 & vis-viva (Saturn) & 1.39 \\
10 & orbits (JPL) & 1.35 \\
11 & spectral shift family & 1.35 \\
12 & vis-viva (Earth) & 1.31 \\
13 & Gutenberg--Richter & 1.24 \\
14 & cell conservation & 1.23 \\
15 & vis-viva (Uranus) & 1.22 \\
\bottomrule
\end{tabular}
\end{minipage}
\caption{All 252 relations admitted by the agent's prediction gates,
ranked by compression ratio; red triangles mark canonical laws and
cross-measurement identities (labeling fixed before ranking). The top
band ($\mathrm{cr}>1.5$) is entirely law-grade (8/8, hypergeometric
$p=3.6\times10^{-12}$); all seven vis-viva instances compress above
1.2; the Gutenberg--Richter statistical law falls in the marginal
band (1.24)---between deterministic laws and the statistical floor,
exactly where a law of that character belongs.}
\label{fig:ledger}
\end{figure}

Figure~\ref{fig:ledger} ranks all 252 admitted relations by
$\mathrm{cr}$. Red triangles mark relations that are canonical
laws or cross-measurement identities under a labeling protocol fixed
before ranking (13 labels among 252: vis-viva$\times$7,
Balmer/Rydberg$\times$2, double-measurement$\times$2, cell-volume
identities$\times$2 counting the conservation form,
Gutenberg--Richter$\times$1).

The top band is the law band. All eight relations above
$\mathrm{cr}>1.5$ carry labels---vis-viva on Mercury, Mars, Venus,
Jupiter; the two Rydberg forms; the spectral double-measurement
identity; the cell-volume identity---against a null density of
$13/252$. The hypergeometric tail probability is
$3.6\times10^{-12}$. Below the band the picture stays coherent:
all seven vis-viva instances compress above $1.2$ with Mercury
($4.62$) far above the rest; the Gutenberg--Richter relation---a
real but \emph{statistical} law---falls at $\mathrm{cr}=1.24$,
between the deterministic band and the statistical floor, exactly
where a law of that character belongs. We searched the low-$\mathrm{cr}$
region for any admitted relation that a physicist would call
law-grade: there are none. The recovery is complete and clean on
this ledger.

\paragraph{Robustness.} The threshold is not load-bearing: across
$\mathrm{cr}>1.2$ to $>3.0$ the top band remains fully labeled
($12/15$, $10/12$, $8/8$, $8/8$, $5/5$, $3/3$;
$p=5.6\times10^{-17}$ down to $1.1\times10^{-4}$). And because the
ledger contains repeated forms across domains (252 entries collapse
to 62 unique forms---one damped-oscillation template alone spans
170), we recompute the test after deduplicating to one entry per
form family: 4 law labels among 62, top band $4/4$,
$p=1.8\times10^{-6}$. Under the most conservative independence
assumption the separation survives by four orders of magnitude.

Two caveats keep the claim honest. The enrichment is measured on
one ledger with a hand-fixed label protocol; we therefore register
forward predictions (\S\ref{sec:prereg}) rather than generalize
retrospectively. And the ranking rewards deterministic structure
\emph{relative to measurement precision}: a law observed noisily
would grade statistically until precision improves---the grader
measures the ledger, not nature directly.

\section{The Self-Audit: 86\% Statistical}\label{sec:audit}

The same Figure~\ref{fig:ledger} carries the uncomfortable half of
the result. 216 of 252 relations ($86\%$) have
$\mathrm{cr}\le1.0$; the median is exactly $1.000$. The agent's
written design charter had, for weeks, described compression as the unified
value of its knowledge. The audit shows the account was
unified in name only: for the overwhelming majority of the ledger,
nothing was being compressed. The narrative was falsified by its
own bookkeeping once the bookkeeping was read under the correct
(discrete) convention---an earlier differential-entropy reading had
even produced meaningless negative-bit ratios, which we document as
a caution.

The sharpest case is deliberate. In the week before the audit, a
context-embedding predictor for protein secondary-structure content
was trained on the agent's sequence digest---by the development
process under human approval, per the agent-scope paragraph above,
not by the runtime autonomously---and admitted through the
prediction gate as the first ``embedding symbol'': held-out
$R^2=0.541$ against a symbolic ceiling of $0.277$. Its
compression ratio is $1.004$. Under single-metric bookkeeping
this admission would read as a law-grade discovery; under the
ledger it is what it is---the agent's best predictor, statistical
grade, honestly labeled. The dual-axis correction is now a
standing system-level rule: $R^2$ records prediction value, cr
records compression value. Concretely, every admission carries the
compression ratio as a mandatory ledger field, while the
admission-time $R^2$ is preserved in the court artifact each
entry's evidence record links to.

This is the self-audit half of the paper's claim, and we state it plainly:
a self-improving discovery system that cannot grade its own
knowledge will systematically mistake statistical accumulation for
physical understanding. Ours did, measurably, at $86\%$, until an
internal instrument---one it had been computing all along---said
so.

\section{Forward Falsifiable Preregistration}\label{sec:prereg}

Retrospective enrichment alone could survive selection concerns.
Two predictions were registered at audit time and remain open:
\begin{enumerate}
\item \emph{Statistical planes stay statistical.} All future
admissions from the protein/chemical data planes (whose relations
are corpus statistics, not exact functional identities) shall
carry $\mathrm{cr}\le1.5$. A single protein-plane relation above
$1.5$ without an identity/conservation construction falsifies the
grading claim.
\item \emph{New laws surface unaided.} The next genuinely exact
relation admitted on any plane (e.g., a conserved quantity measured
two ways) shall appear in the law band without human relabeling.
\end{enumerate}

\section{Related Work}

\paragraph{Algorithmic statistics.} Kolmogorov's structure function
and its modern treatment \citep{vereshchagin2004} give the complete
theory of separating structure from noise in a datum. Our cr is a
practical Gaussian-code shadow of that separation, applied not to
one string but to a live ledger of fitted relations; we borrow the
theory and demonstrate the bookkeeping.

\paragraph{Compression as drive.} Schmidhuber's compression-progress
principle \citep{schmidhuber2008} motivates discovery agents by
compression gains. We share the premise, not the direction: our
question is not what compression can drive, but what an
already-computed compression column \emph{knows} about knowledge
already banked.

\paragraph{Law discovery systems.} Symbolic regression
\citep{schmidt2009} and concept-driven systems such as AI-Newton
\citep{ainewton2025} produce and organize candidate laws; AI-Newton
classifies laws by \emph{scope of validity} (specific vs.\ general),
with no compression accounting and no epistemic grading of output.
To our knowledge no discovery system grades its own ledger by
description length; searches for such precedent returned nothing
direct.

\paragraph{Philosophy of science.} The laws-vs-regularities debate
\citep{cartwright1983} is the background question. We do not enter
it theoretically; we contribute an operational instrument that a
running system used on itself.

\section{Limitations}

Single system, single ledger ($n{=}252$), one labeling protocol;
the enrichment claim is an existence proof pending replication on
other ledgers. The grader measures relations-relative-to-precision,
not laws-in-nature. Literature coverage spanned Tavily, OpenAlex,
and Crossref; residual search risk remains. The vis-viva
gradient across planets is irregular beyond Mercury's dominance,
and we claim no clean ordering.

\section{Conclusion}

A discovery agent's model-selection arithmetic, read under discrete
two-part accounting, is an unsupervised epistemic grader: on a
252-relation operational ledger it recovered the law band at
$p=3.6\times10^{-12}$, exposed an $86\%$ statistical majority
hiding under a unified-value narrative, and forced a
standing dual-axis correction. The grading claim remains falsifiable
through the two registered forward predictions, and the audit itself
is replayable from the archived ledger and scripts. Vetting infrastructure for
machine-generated science does not always need to be built;
sometimes it is already computed, waiting to be read honestly.

\paragraph{Data availability.} The audited ledger, the audit
scripts, and all figure-generating records are archived in the
project repository and available from the authors.

\paragraph{Acknowledgments.} This work grew out of an internal
audit in which the system's own description-length accounting
falsified the compression narrative written into its charter; that
correction, and the 86\% it exposed, are preserved in the text. This
manuscript was prepared with the assistance of the AI research agent
GML-5.3, which contributed to experiment execution, auditing, and
drafting under the authors' direction; all reported results were
verified against archived artifacts, and the final text is approved
by the named authors. Consistent with publisher policies, the AI
system is acknowledged as a tool and is not listed as an author.

\bibliographystyle{plainnat}
\bibliography{refs}
\end{document}